\pdfoutput=1

\documentclass[11pt]{article}

\usepackage[preprint]{acl}

\usepackage{times}
\usepackage{latexsym}
\usepackage[whole]{bxcjkjatype}
\usepackage[T1]{fontenc}

\usepackage[utf8]{inputenc}

\usepackage{microtype}

\usepackage{inconsolata}

\usepackage{ascmac}
\usepackage{graphicx}
\usepackage{multirow}
\usepackage {nicematrix}
\usepackage{booktabs}

\usepackage{color}
\definecolor{ruled}{RGB}{192,192,192}

\title{
An Empirical Analysis of Factual Errors in Human-Written Text and\\Its Application to Factual Error Detection
}

\author{
    Kazuma Iwamoto \\
    \And
    Kazumasa Omura \\
    Nikkei Inc. \\
    \texttt{\{kazuma.iwamoto, kazumasa.omura, shotaro.ishihara\}@nex.nikkei.com} \And
    Shotaro Ishihara\thanks{This work was conducted while the author was with Nikkei Inc.}
}

\begin{document}
\maketitle
\begin{abstract}
Factual Error Detection (FED), which is the task of identifying factually incorrect spans in a given text, has long been recognized as an important research problem.
However, with the rapid rise of large language models (LLMs), research attention has shifted toward factual errors specific to LLM-generated text (hallucinations) and their detection.
As a result, the detection of factual errors in human-written text has been relatively neglected.
To address this gap, we first distill a taxonomy of human-induced factual errors by analyzing corrections of newspaper articles, a representative source of text that is guaranteed to be human-written and contains few grammatical errors.
Our analysis revealed that there are characteristic categories such as kanji misconversions and unit errors, which are not focused in existing hallucination benchmarks.
Based on the taxonomy, we then evaluate the FED capability of vanilla LLMs on synthesized realistic test cases and real corrections.
Experimental results demonstrated that even high-performance LLMs such as GPT-5.4 achieved only word-level F1 score of 52\% on the synthetic evaluation data, highlighting the task difficulty.
Furthermore, a detailed analysis by detection difficulty revealed the current state of FED.
\end{abstract}

\section{Introduction}
\label{sec:intro}

\begin{figure}[!t]
    \centering
    \includegraphics[width=\linewidth]{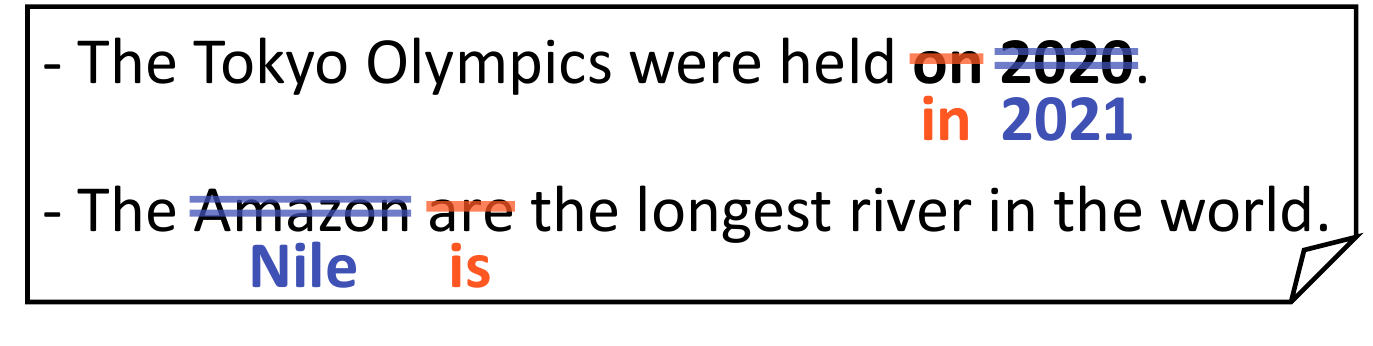}
    \caption{
        Difference between proofreading (orange line) and revising (indigo double line).
        Proofreading typically refers to correcting grammatical errors, whereas revising involves identifying and rectifying factual inaccuracies (text spans that are not based on facts) in text.
    }
    \label{fig:proofreading-and-revising}
\end{figure}

In the writing process, two key stages ensure the quality of text: \emph{proofreading} and \emph{revising}~\cite{du-etal-2022-understanding-iterative}.
Proofreading addresses grammatical errors, whereas revising addresses factual errors (text spans that are not based on facts), as illustrated in Figure~\ref{fig:proofreading-and-revising}.\footnote{
Proofreading might also address factual errors, but we distinguish between proofreading and revising in this sense.
}
These stages are highly specialized and painstaking because they require extensive linguistic and world knowledge to detect and correct errors in text.
Therefore, there is a strong demand for systems that support proofreading and revising.

This study focuses on revising rather than proofreading.
There have been numerous initiatives toward proofreading support~\cite{bryant-etal-2023-grammatical}, whereas those toward revision support remain scarce~\cite{chen-etal-2023-converge}.
However, recent advances in large language models (LLMs) make them a natural candidate for revision support.
Since LLMs have acquired extensive world knowledge through pre-training on large-scale raw corpora~\cite{petroni-etal-2019-language,roberts-etal-2020-much,allen-li-2024-physics}, we hypothesize that they are suitable for revising even without external knowledge grounding.

\begin{figure*}[!t]
    \centering
    \includegraphics[width=\linewidth]{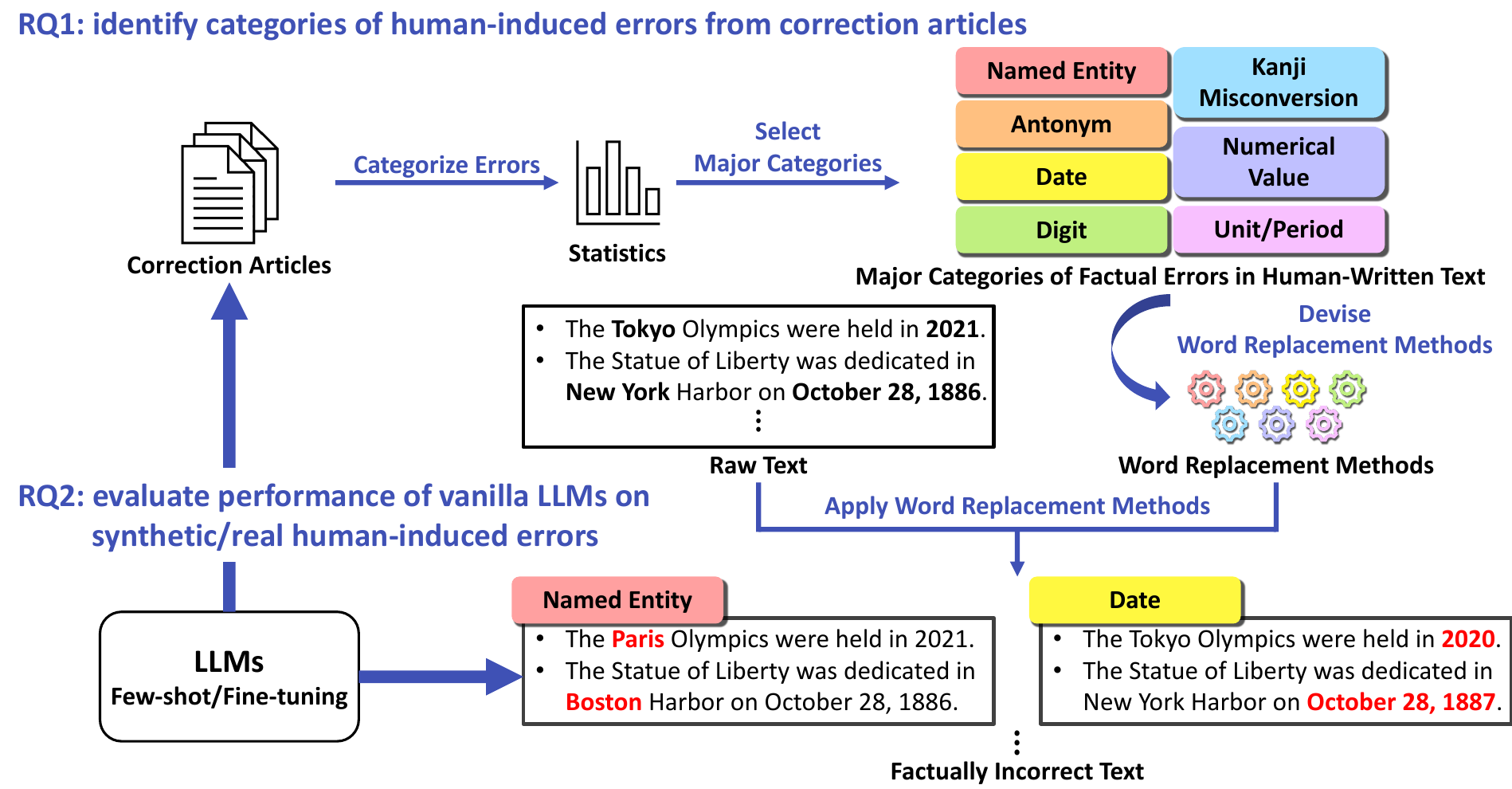}
    \caption{
        Overview of this study.
        First, we categorize errors reported in correction articles.
        We next devise word replacement methods for simulating the factual errors.
        We then apply the word replacement methods to raw text and automatically generate factually incorrect text.
        Finally, we evaluate the performance of LLMs on FED using the factually incorrect text and the original correction articles.
    }
    \label{fig:overview}
\end{figure*}

Revising comprises two sub-stages: detecting text spans that require revision, and correcting them based on facts.
We adopt the former, Factual Error Detection (FED), as our target task, although both stages remain challenging~\cite{he-etal-2024-improving-factual}.
This task is crucial for making accurate corrections and indispensable for supporting editorial work where humans have the final authority over corrections.
Particularly, overlooking factual errors in newspaper articles and educational materials, which carry a profound social impact, can lead to serious issues.
Therefore, automatic detection of factual errors in human-written text is desired for revision support.

Nevertheless, recent research trends have not adequately addressed the analysis and detection of factual errors in human-written text~\cite{huang-etal-2024-survey}.
Since the advent of LLMs, research has increasingly focused on factual errors specific to LLM-generated text, known as \emph{hallucinations}.
Although revision support should target human-written text, there has been little research that directly addresses human-induced factual errors.

To bridge this research gap, we aim to analyze factual errors in human-written text and evaluate the capability of LLMs to detect them.
Specifically, we set the following research questions (RQs):
\begin{description}
    \item[RQ1] What types of factual errors occur in human-written text?
    \item[RQ2] To what extent can vanilla LLMs detect factual errors in human-written text?
\end{description}

Regarding RQ1, it is not obvious which language resources can be used for the analysis.
For instance, an edit history of Wikipedia contains edits addressing not only factual errors but also other types of errors, and is therefore noisy.
In this study, we utilize \emph{correction articles},\footnote{
Examples can be seen at list of corrections for articles in \emph{New York Times} (\url{https://www.nytimes.com/international/section/corrections}).
\label{footnote:correction_articles}
} which report errors and their corrections of already-published newspaper articles.
Since newspaper articles are guaranteed to be written by humans\footnote{
The AI usage policy of Nikkei, of which articles we utilized, can be available at \url{https://asia.nikkei.com/announcements/nikkei-s-view-on-ai-in-journalism?n_cid=DSBNNAR}.
} and well-proofread (i.e., contain few grammatical errors), errors reported in correction articles can be regarded as representative of factual errors in human-written text.

Regarding RQ2, the lack of datasets focusing on factual errors in human-written text is problematic.
Although correction articles are regarded as real test cases, their size is limited.
We therefore utilize synthetic data that simulates human-induced factual errors based on the analysis of correction articles.
Automatic generation can create a broad range of test cases, thereby enabling the construction of a robust dataset for evaluating NLP models~\cite{ribeiro-etal-2020-beyond}.
Since we hypothesize that vanilla LLMs are suitable for revising, we investigate it through a detailed analysis of model performance across error categories and difficulties.

\begin{table*}[!t]
    \centering
    \small
    \begin{NiceTabular}{lccl}
        \toprule
            \textbf{Task} &
            \textbf{Span-level} &
            \textbf{Human-induced Errors} &
            \textbf{Detection vs. Correction} \\
        \midrule
            Fact Verification &
            $\times$ &
            $\sim$ &
            Detection \\
            Factual Error Correction (FEC) &
            $\surd$ &
            $\sim$ &
            Correction \\
            Hallucination Detection &
            $\surd$ &
            $\times$ &
            Detection \\
        \midrule
            \textbf{This work: FED for Human-Written Text} &
            $\surd$ &
            $\surd$ &
            Detection \\
        \bottomrule
    \end{NiceTabular}
    \caption{
        Comparison of related tasks from three perspectives:
        (i) whether the task operates at the \textit{span level},
        (ii) whether the target is \textit{human-induced} factual errors, and
        (iii) whether the main objective is \textit{detection} or \textit{correction}.
        Our work uniquely focuses on span-level detection of human-induced factual errors.
        ``$\sim$'' denotes that the target is factual errors intentionally made by humans.
    }
    \label{tab:taxonomy}
\end{table*}

The overview of this study is illustrated in Figure \ref{fig:overview}.
Our contributions are summarized as follows:
\begin{itemize}
    \item We identified categories of factual errors that commonly occur in human-written text by analyzing correction articles.
    \item We evaluated FED performance of powerful vanilla LLMs on synthetic and real test cases and revealed what human-induced factual errors are challenging to detect.\footnote{
    We released reproduction code at \url{https://github.com/Nikkei/fed4human}.
    \label{footnote:github}
    }
\end{itemize}

\section{Related Work}
\label{sec:related_work}

Tasks regarding detecting factual errors in text include fact verification, factual error correction, and hallucination detection.
As shown in Table~\ref{tab:taxonomy}, this section outlines each task and describes how it differs from our work.

\paragraph{Fact Verification}

Fact Verification, also called fact-checking, is the task of judging whether given text presents misinformation.
One widely used benchmark for this task is FEVER~\cite{thorne-etal-2018-fever}, which formulates fact verification as a three-way classification of a given claim sentence into \textsc{Supported}, \textsc{Refuted}, or \textsc{NotEnoughInfo} with reference to evidence.
This benchmark comprises 185,445 claim sentences based on Wikipedia, each annotated with a label and evidence.
However, it does not consider which part of each sentence is factually incorrect.
As exemplified by FEVER, fact verification often focuses on judging the factual correctness of an entire given claim sentence based on retrieved evidence~\cite{augenstein-etal-2019-multifc,wadden-etal-2020-fact,kotonya-toni-2020-explainable-automated,jiang-etal-2020-hover,ma-etal-2024-ex}, where retrieval performance matters.

\paragraph{Factual Error Correction}

Factual Error Correction (FEC) is the task of minimally editing given text to ensure factual correctness.
A major dataset for this task is \textsc{FecData}~\cite{thorne-vlachos-2021-evidence}, which utilizes the intermediate annotations of FEVER.
Specifically, since FEVER has been built by manually modifying sentences sampled from Wikipedia to factual and nonfactual ones, \textsc{FecData} reuses the pairs of sentences before and after the modification for FEC.
However, this dataset fails to address factual errors that are likely to occur in human-written text as humans intentionally made sentences factually incorrect.
It also does not consider categories of factual errors, making detailed analysis impractical.

Regarding factual error detection, a few studies have explored it as an upstream stage to FEC.
For instance, \citet{thorne-vlachos-2021-evidence} and \citet{he-etal-2024-improving-factual} selected tokens in a given claim sentence that were either randomly sampled or absent in retrieved evidence as candidate errors.
\citet{fatahi-bayat-etal-2023-fleek} extracted facts from a given claim sentence in the subject–predicate–object triple format and regarded the entities as candidate errors.
These studies have focused primarily on correcting factual errors; the preceding sub-stage of detecting factually incorrect spans in human-written text has not been explored.

\begin{table*}[!t]
    \centering
    \small
    \begin{NiceTabular}{c|c|l|c}
        \toprule
            Supercategory &
            Category &
            \multicolumn{1}{c|}{Example of Correction} &
            \multicolumn{1}{c}{Count} \\
        \midrule
            \Block{5-1}{Word Error} & 
            \raisebox{-0.3125\height}{\shortstack{Named\\Entity}} &
            \begin{minipage}[h]{0.6\textwidth}
                「ワシントン州」は「オレゴン州」の誤りでした\\
                (it was not ``Washington State'' but ``Oregon State'')
            \end{minipage} &
            57 \\
        \cmidrule{2-4}
             &
            \raisebox{-0.3125\height}{\shortstack{Kanji\\Misconversion$^{\dagger}$}} & 
            \begin{minipage}[h]{0.6\textwidth}
                「本田英明部長」は「本多英明部長」の誤りでした\\
                (it was not ``Hideaki Honda, manager'' but ``Hideaki Honda, manager'')
            \end{minipage} &
            23 \\
        \cmidrule{2-4}
            &
            Synonym & 
            \begin{minipage}[h]{0.6\linewidth}
                「売上高」は「営業利益」の誤りでした\\
                (it was not ``sales'' but ``operating profit'')
            \end{minipage} &
            18 \\
        \cmidrule{2-4}
             &
            Antonym & 
            \begin{minipage}[h]{0.6\linewidth}
                「若年就業率が高い」は「若者失業率」の誤りでした\\
                (it was not ``youth employment rate so high'' but ``youth unemployment rate'')
            \end{minipage} &
            14 \\
        \cmidrule{2-4}
             &
            Other & 
            \begin{minipage}[h]{0.6\textwidth}
                「予算案を議会に提案した」は「提案する見込み」の誤りでした\\
                (it was not ``proposed a budget to Congress'' but ``prospect to propose'')
            \end{minipage} &
            9 \\
        \midrule
            \Block{4-1}{Number Error} &
            \raisebox{-0.3125\height}{\shortstack{Numerical\\Value}} &
            \begin{minipage}[h]{0.6\textwidth}
                「資本金3000万円」は「3500万円」の誤りでした\\
                (it was not ``capital of 30 million yen'' but ``35 million yen'')
            \end{minipage} &
            29 \\
        \cmidrule{2-4}
             &
            Date & 
            \begin{minipage}[h]{0.6\textwidth}
                「2020年6月」は「21年6月」の誤りでした\\
                (it was not ``June 2020'' but ``June 2021'')
            \end{minipage} & 
            24 \\
        \cmidrule{2-4}
             &
            Digit$^{\dagger}$ & 
            \begin{minipage}[h]{0.6\textwidth}
                「2トン」は「2万トン」の誤りでした\\
                (it was not ``two tons'' but ``20 thousand tons'')
            \end{minipage} &
            15 \\
        \cmidrule{2-4}
             &
            Unit/Period$^{\dagger}$ & 
            \begin{minipage}[h]{0.6\textwidth}
                「1バレルあたり」は「1ガロンあたり」の誤りでした\\
                (it was not ``per barrel'' but ``per gallon'')
            \end{minipage} &
            11 \\
        \midrule
        \midrule
            \Block{1-2}{Grammatical Error} & & 
            \begin{minipage}[h]{0.6\linewidth}
                「11日1日」は「11月1日」の誤りでした\\
                (it was not ``11th 1st'' but ``November 1st'')
            \end{minipage} &
            4 \\
        \midrule
            \Block{1-2}{Phrase-Level Error} & & 
            \begin{minipage}[h]{0.6\textwidth}
                「デロイトに制裁を科した」は「KPMGを責めた」の誤りでした\\
                (it was not ``imposed sanctions on Deloitte'' but ``criticized KPMG'')
            \end{minipage} &
            30 \\
        \bottomrule
    \end{NiceTabular}
    \caption{
        Classification result of factual errors and their corrections reported in 234 Japanese correction articles.
        $\dagger$ denotes the category of factual errors that is characteristic of human-written text.
        We focus on ``word error'' and ``number error'' considering that they occur frequently and can be simulated by word replacement.
    }
    \label{tab:category}
\end{table*}

\paragraph{Hallucination Detection}

Hallucination detection literally aims to detect hallucinations, factually incorrect spans in LLM-generated text.
Growing concern about hallucination has prompted the development of several benchmarks for hallucination detection~\cite{huang-etal-2024-survey}.
One example is HaluEval~\cite{li-etal-2023-halueval}, which consists of 35k hallucinated responses by ChatGPT\footnote{\url{https://chatgpt.com/}\label{footnote:chatgpt}} to user queries.
It has been extended to HaluEval 2.0~\cite{li-etal-2024-dawn}, where hallucinations are classified into six types (Entity-error, Relation-error, Incompleteness, Outdatedness, Overclaim, and Unverifiability).
\citet{mishra-etal-2024-finegrained} also classified hallucinations into six types (Entity, Relation, Sentence, Invented, Subjective, and Unverifiable) and built 1k passages annotated with hallucination type and factually incorrect span.
However, this task by definition targets factual errors in LLM-generated text and may overlook those in human-written text.

\section{Taxonomy of Human-Induced Factual Errors}

Section~\ref{sec:related_work} indicates that related work has not clarified categories of factual errors that commonly occur in human‑written text nor investigated the capability of LLMs to detect them.
As a first step, we investigate human-induced factual errors.

\subsection{Analysis of Correction Articles}
\label{subsec:analysis}

As described in Section~\ref{sec:intro}, we utilized correction articles\footref{footnote:correction_articles} for analyzing categories of factual errors that commonly occur in human-written text.
Specifically, we collected 234 correction articles of the Nikkei Morning and Evening Editions\footnote{\url{https://www.nikkei.com/}\label{foot:nikkei}} published \emph{between January 2020 and December 2023} and manually categorized the errors reported in them.

The categorization was performed by the first author. 
Subsequently, the classification categories were established and refined through collaborative discussions among all the authors referring to actual correction articles.

The classification result is organized in Table~\ref{tab:category}.
We classified factual errors in human-written text into 11 categories.
For space limitation, we present the definitions of categories that appear to require further clarification below.
\begin{description}
    \item [Kanji Misconversion] Corrections of homophone errors that arise when an incorrect kanji with the same reading is selected during conversion from hiragana to kanji.
    \item [Numerical Value] Corrections of number errors to the extent that digits remain the same.
    \item [Digit] Corrections of number errors involving digit changes.
    \item [Phrase-Level Error] Corrections of phrase-level errors that cannot be fixed by word replacement alone.
\end{description}
All the definitions of categories are included in Appendix~\ref{appendix:definitions_of_categories}.

As for supercategories, there were 121 word errors, 79 number errors, and 30 phrase-level errors, respectively.
Numerous corrections involve only a single word or number, which demonstrates that a considerable portion can be reproduced with simple word replacement.

\subsection{Comparison with Hallucination Benchmarks}

To analyze human-specific factual errors, we classified factual errors included in hallucination benchmarks according to the taxonomy in Table \ref{tab:category}.
Specifically, we examined 245 hallucinated responses to questions labeled ``knowledge-based'' in JTruthfulQA,\footnote{\url{https://github.com/nlp-waseda/JTruthfulQA}} the Japanese version of TruthfulQA~\cite{lin-etal-2022-truthfulqa}.
JTruthfulQA is built by collecting LLM responses to questions and annotating their correctness by humans;
therefore, it is a collection of hallucinations that naturally occur.
We further examined 375 hallucinated responses to questions where their answers are numbers in HaluEval~\cite{li-etal-2023-halueval} to augment number errors.
Although HaluEval contains hallucinations that are intentionally induced, we utilized them as a supplementary resource.

As a result, we found that only one case falls into the Kanji Misconversion, Digit, and Unit/Period categories.
Most numerical hallucinations in HaluEval involved number errors where the correct numbers are mistaken for different ones with the same number of digits.
In JTruthfulQA, there were numerous hallucinations where named entities are mistakenly replaced with other ones of the same type (e.g., Shinzo Abe and another previous prime minister of Japan).
Thus, we conclude that these categories are characteristic of human-written text.

\section{Synthetic Dataset Construction}
\label{sec:synthetic}

In this section, we construct a synthetic dataset that simulates factual errors in human-written text, as one application of the taxonomy derived from correction articles.
The procedure for data synthesis consists of the following steps (cf. Figure~\ref{fig:overview}).
\begin{enumerate}
    \item Devise word replacement methods for simulating the factual errors of major categories.
    \item Generate factually incorrect text by applying the devised methods to raw text.
\end{enumerate}
The following subsections explicate each step.

\subsection{Devising Word Replacement Methods}
\label{subsec:method}

Based on our analysis in Section~\ref{subsec:analysis}, we devised word replacement methods for simulating the common patterns of factual errors in human-written text.
In this study, we adopted seven categories (Named Entity, Kanji Misconversion, Antonym, Numerical Value, Date, Digit, and Unit/Period)\footnote{
We excluded Synonym because we often failed to make a factually incorrect text by word replacement, thereby degrading the quality of a dataset.
} for evaluation, considering both frequency and ease of simulation.
The following paragraphs describe how to select a word or number to be replaced (hereafter, \textbf{target word}) and generate its replacements for each category.
Note that we replace every target word in text to ensure contextual consistency.

\paragraph{Named Entity}

We focus on five named entity categories that commonly appear in correction articles: names of persons, organizations, job titles, locations, and countries.
First, we extract a named entity that falls into one of the aforementioned categories for a target word.
We then mask it and generate candidates of its replacements using a text generation model, T5~\cite{raffel-etal-2020-exploring}, which can complete a masked span with multiple tokens.
Finally, we apply named entity recognition to each candidate and retain it as a replacement if it belongs to the same category as the target word.
We used GiNZA\footnote{
\url{https://github.com/megagonlabs/ginza} (the \emph{ja\_ginza\_bert\_large} model)
\label{footnote:ginza}
} for named entity recognition and $\text{T5}_{\text{XL}}$ pretrained on a Japanese portion of mC4~\cite{xue-etal-2021-mt5} and Wikipedia\footnote{\url{https://huggingface.co/retrieva-jp/t5-xl}} for candidate generation.

\paragraph{Kanji Misconversion}

We target kanji misconversions of personal names because they were the most frequent among the kanji misconversions in correction articles.
Thus, a kanji compound word in an extracted personal name is replaced with another one that has the same reading.
Specifically, a kanji compound word is transliterated into kana and 
reconverted into another one to simulate kanji misconversions. 
We used GiNZA\footref{footnote:ginza} for extracting personal
names, pykakasi\footnote{\url{https://codeberg.org/miurahr/pykakasi}} for kanji-kana conversion,
and mozcpy\footnote{\url{https://github.com/ikegami-yukino/mozcpy}} for kana-kanji conversion.

\paragraph{Antonym}

We regard a verbal noun as a target word and generate candidates of its antonyms using word2vec~\cite{mikolov-etal-2013-distributed} and an LLM.
Specifically, we first use word2vec to select the five words most similar to a target word for candidates, leveraging its characteristic that assigns high similarity not only to synonyms but also to antonyms~\cite{mrksic-etal-2016-counter}.
We then prompt an LLM to generate candidates and adopt the intersection with those selected by word2vec for replacements.
We used MeCab\footnote{\url{https://taku910.github.io/mecab/}\label{footnote:mecab}} for extracting verbal nouns, word2vec trained on Japanese Wikipedia\footnote{\url{https://www.cl.ecei.tohoku.ac.jp/~m-suzuki/jawiki_vector/}} and the 8B instruction-tuned Llama 3.1 Swallow~\cite{fujii-etal-2024-continual}\footnote{
\url{https://huggingface.co/tokyotech-llm/Llama-3.1-Swallow-8B-Instruct-v0.1}
\label{footnote:swallow}} for candidate generation.

\paragraph{Numerical Value}

We extract a number using a regular expression and replace it with a random number within the same number of digits.

\paragraph{Date}

We substitute the year, month, or day component of a date in text with another valid value.
When replacing a year component, we choose a random value within $\pm{5}$ years of the original to avoid making text unrealistic.

\paragraph{Digit}

We select a numerical expression for a target word when it satisfies either of the following conditions: (1) it ends in two or more zeros, or (2) it combines Arabic numerals with Japanese numeral units such as ``16億 (1.6 billion)''.
An expression meeting (1) is replaced with a scaled‑down or scaled‑up numeral, whereas that meeting (2) is replaced with another Japanese numeral unit.

\paragraph{Unit/Period}

We extract a numeral classifier for a target word and substitute it with another similar word using word2vec.
We used MeCab\footref{footnote:mecab} for extracting numeral classifiers and the same word2vec model as that used for the Antonym category.

\subsection{Generating Synthetic Data}
\label{subsec:dataset}

We synthesized training and evaluation data by applying the word replacement methods described in Section~\ref{subsec:method} to raw text.
Examples are provided in Appendix~\ref{appendix:example_data}.

\paragraph{LLM-based Filtering}

When generating replacements for the Named Entity and Antonym categories, a replacement may be an absolute synonym of its target word such as ``欧州 (Europe)'' and ``ヨーロッパ (Europe)''.
To reduce such noise, we apply an LLM-based filtering to replacements for these categories before building training and evaluation data.
Specifically, we instruct the 8B instruction-tuned Llama 3.1 Swallow\footref{footnote:swallow} to judge whether a target word and its replacement are absolute synonyms.
The prompt template for this filtering is included in Appendix~\ref{appendix:template}.
As a result, we filtered out approximately 20\% of the generated data.

\paragraph{Evaluation Data}

We employed newspaper articles published by Nikkei Inc.\footref{foot:nikkei} \emph{in December 2020} for raw text.
We generated test cases and randomly sampled 100 of them for each adopted category.
Regarding the Named Entity category, we sampled 100 test cases for each named entity category (i.e., names of persons, organizations, job titles, locations, and countries) to enhance diversity.
We also included 100 test cases with no word replacement.
Owing to the automatic generation, some noise might be included in the evaluation data.
We discuss the proportion of noise in Section~\ref{sec:basis}.

\paragraph{Training Data}

We generated training cases from Nikkei articles published \emph{between January 2015 and December 2019} and randomly sampled up to 5,000 for each adopted category.
As with the evaluation data, we sampled training cases for each named entity category.
22,053 negative training cases, articles with no word replacement, were also included in the training data.

\begin{table}[!t]
    \small
    \centering
    \begin{NiceTabular}{l|r|r}
        \toprule
            \multicolumn{1}{c|}{Category} & \multicolumn{1}{c|}{Train} & \multicolumn{1}{c}{Eval} \\
        \midrule
            Named Entity & 17,329 & 500\\ 
            Kanji Misconversion & 5,000 & 100\\
            Antonym & 5,000 & 100\\
            Numerical Value & 5,000 & 100\\ 
            Date & 5,000 & 100  \\ 
            Digit & 5,000 & 100 \\ 
            Unit/Period & 5,000 & 100 \\ 
            no word replacement & 22,053 & 100 \\
        \bottomrule
    \end{NiceTabular}
    \caption{Statistics of the constructed synthetic dataset.}
    \label{tab:dataset}
\end{table}

\begin{table*}[!t]
    \small
    \centering
    \begin{NiceTabular}{l|l|rrr|rrr}    
        \toprule
            \multicolumn{1}{c}{\multirow{2}{*}{Model}} &
            \multicolumn{1}{c}{\multirow{2}{*}{Setting}} &
            \multicolumn{3}{c}{Sentence-Level} &
            \multicolumn{3}{c}{Word-Level} \\
            &
            &
            \multicolumn{1}{c}{Prec.} &
            \multicolumn{1}{c}{Rec.} &
            \multicolumn{1}{c}{F1} &
            \multicolumn{1}{c}{Prec.} &
            \multicolumn{1}{c}{Rec.} &
            \multicolumn{1}{c}{F1} \\
        \midrule
            GPT-5.4 &
            Few-Shot ($k = 6$) &
            \textbf{41.7} &
            46.1 &
            \textbf{43.8} &
            \textbf{78.7} &
            \textbf{38.8} &
            \textbf{52.0} \\
        \midrule
            \Block{2-1}{Qwen3-Swallow-8B} &
            Few-Shot ($k = 6$) &
            9.0 &
            \textbf{48.5} &
            15.2 &
            45.0 &
            6.2 &
            10.9 \\
            &
            Fine-Tuning (QLoRA) &
            41.3 &
            36.4 &
            38.7 &
            49.2 &
            28.7 &
            36.2 \\
        \midrule
            \Block{2-1}{Gemma-4-8B} &
            Few-Shot ($k = 6$) &
            18.4 &
            15.5 &
            16.8 &
            51.6 &
            8.6 &
            14.7 \\
            &
            Fine-Tuning (QLoRA) &
            32.1 &
            25.9 &
            28.7 &
            56.7 &
            21.1 &
            30.8 \\
        \bottomrule
    \end{NiceTabular}
    \caption{
        Performance of detecting factually incorrect spans on the evaluation data.
        Although we also attempted full-parameter fine-tuning, it yielded lower performance than QLoRA;
        therefore, we report only the results of fine-tuning using QLoRA in the main text.
        See Appendix~\ref{sec:full_experiments_nikkei} for full experimental results.
    }
    \label{tab:experimental_results}
\end{table*}

\section{Experiments}
\label{sec:experiments}

We conducted experiments to evaluate the performance of LLMs on the constructed synthetic dataset and real correction articles.
Since our motivation is to investigate the capability of vanilla LLMs as described in Section~\ref{sec:intro}, we focus on frameworks independent of external knowledge such as in-context learning and fine-tuning.

These experiments use proprietary data (Nikkei articles), meanwhile we also constructed a dataset using Japanese articles published on the open-source Wikinews platform\footnote{\url{https://www.wikinews.org/}\label{footnote:wikinews}} for reproducibility and confirmed that the overall performance trends were similar.
The details are described in Appendix~\ref{sec:experiments-wikinews}.

\subsection{Experimental Settings}
We evaluated how well LLMs can detect factually incorrect spans under two settings: sentence-level and word-level.
In the sentence-level setting, we prompt LLMs to excerpt sentences that contain factually incorrect spans.
We treat substrings obtained by splitting text at a punctuation mark as sentences.
In word-level, we instruct LLMs to identify words that correspond to factually incorrect spans by enclosing them with \emph{<|factual\_error\_start|>} and \emph{<|factual\_error\_end|>} tags.
We measured the detection performance by precision, recall, and F1 score.

We employed GPT-5.4, the 8B instruction-tuned Qwen3 Swallow (hereafter, Qwen3-Swallow-8B),\footnote{\url{https://huggingface.co/tokyotech-llm/Qwen3-Swallow-8B-SFT-v0.2}} and the 8B instruction-tuned Gemma 4 (hereafter, Gemma-4-8B)\footnote{\url{https://huggingface.co/google/gemma-4-E4B-it}} as evaluation targets, all of which exhibit strong performance in Japanese.
We investigated few-shot ($k = 6$)\footnote{We set $k$ to 6 because it was the maximum number of few-shot examples that could be provided to Gemma due to computational constraints at the time. For consistency, we used the same value for other models.} performance of each model and performance of Qwen3-Swallow-8B and Gemma-4-8B fine-tuned using QLoRA~\cite{dettemrs-etal-2023-qlora} on the training data described in Section~\ref{subsec:dataset}.
The details of the hyperparameters are provided in Appendix~\ref{appendix:qlora_hyperparameter}.
For convenience, we refer to the fine-tuned models as Qwen3-Swallow-8B-QL and Gemma-4-8B-QL, respectively.

\subsection{Experimental Results}
\label{subsec:experimental_results}

The evaluation results are organized in Table~\ref{tab:experimental_results}.
As for few-shot performance, GPT-5.4 achieved the best, F1 scores of 43.8\% and 52.0\% in the sentence-level and word-level settings, respectively.
Considering that the performance does not change significantly between both settings, GPT-5.4 is capable of recognizing factually incorrect spans to some extent.
On the other hand, both Qwen3-Swallow-8B and Gemma-4-8B models struggled with this task overall.

Turning our attention to fine-tuning performance, both Qwen3-Swallow-8B-QL and Gemma-4-8B-QL exhibited significant improvement in F1-scores compared to their few-shot performance.
However, the highest F1 score in the word-level setting is still 36.2\% achieved by Qwen3-Swallow-8B-QL, which demonstrates the FED task is quite challenging even for recent high-performance LLMs.

\begin{figure}[!t]
    \centering
    \includegraphics[width=0.85\linewidth]{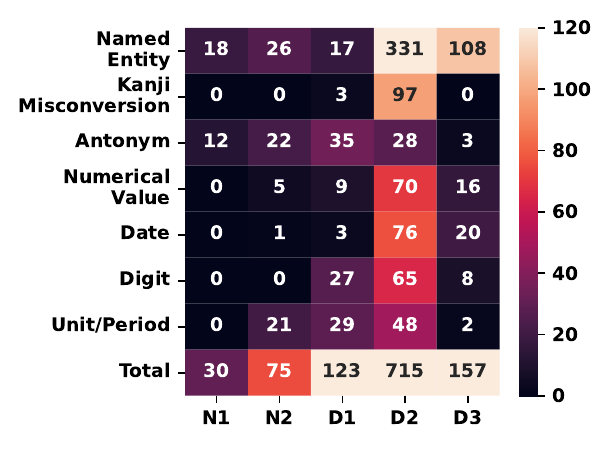}
    \caption{Annotation result of detection difficulties.}
    \label{fig:annotation_result}
\end{figure}

\begin{figure}[!t]
    \centering
    \includegraphics[width=0.85\linewidth]{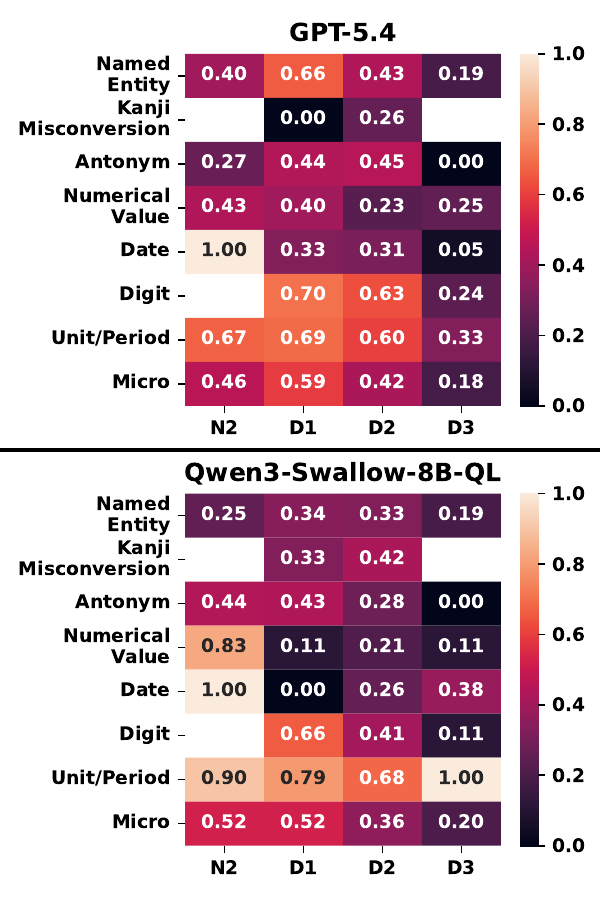}
    \caption{
        F1 scores for each category and detection difficulty.
        The blank cell represents that no case falls into the corresponding category and detection difficulty.
    }
    \label{fig:performance_matrix}
\end{figure}

\subsection{Analysis by Detection Difficulty}
\label{sec:basis}

In the FED task, test cases vary in difficulty: some factual errors can be detected solely from the surrounding context, or others require world knowledge.
Thus, we define detection difficulties and analyze the performance for each category in detail by manually annotating evaluation data with them.
Definitions of detection difficulties are as follows:
\begin{description}
    \item[Noise 1~(N1)] Not contain any factual error.
    \item[Noise 2~(N2)] Detectable through grammatical unnaturalness or solely from a replacement word or number.
    \item[Difficulty 1~(D1)] Detectable solely from the surrounding context.
    \item[Difficulty 2~(D2)] Detectable with world knowledge in addition to the surrounding context.
    \item[Difficulty 3~(D3)] Difficult to detect even with world knowledge.
\end{description}
Examples for each detection difficulty are provided in Appendix~\ref{appendix:examples_for_ddt}.

The annotation result is displayed in Figure~\ref{fig:annotation_result}.
D2 accounts for the majority of the evaluation data, indicating that evaluation data is knowledge-intensive to detect factual errors.
In addition, we can see that a substantial portion of the evaluation data contain factually incorrect spans as the numbers of the test cases that fall into N1 and N2 are just 30 and 75, respectively.
In the Antonym category, D1 appears most frequently, suggesting that factual errors related to antonyms are mainly detectable from their surrounding context.

Figure~\ref{fig:performance_matrix} illustrates F1 score of GPT-5.4 and Qwen3-Swallow-8B-QL, which perform well on the evaluation data, for each category and detection difficulty.
First, we focus on the micro-F1 for each detection difficulty.
The performance for D1 (detectable from the surrounding context)  of each model is the highest among all the detection difficulties.
From this result, we speculate that current LLMs can detect factual errors to a large extent if these can be verified from contextual consistency.
Conversely, the performance for D2 (detectable with world knowledge) is much lower than that for N2 and D1.
This result indicates that the world knowledge possessed by LLMs is still limited or not fully utilized.

Next, we focus on F1 score for each category and detection difficulty.
For both models, the performance was relatively high for the Unit/Period and Digit categories, even for D2.
The probable reason is that a single error in unit/period or digit often leads to a significant deviation from facts.
For instance, if the Guinness World Record for the highest body weight were stated as ``635 g'' instead of ``635 kg'', even someone who does not know the exact weight could detect the error using commonsense knowledge.
Thus, we speculate that these errors can be detected by heuristics.

In addition, the micro-F1 for N2 of Qwen3-Swallow-8B-QL is higher than that of GPT-5.4.
We presume that fine-tuning improves the ability to identify relatively easy-to-detect factual errors.

\subsection{Performance on Real Corrections}

We also created 147 real test cases from 234 correction articles used in Section~\ref{subsec:analysis} by curating errors that are not in figures or tables and can be fixed by word replacement.
Using the data, we evaluated few-shot ($k = 6$) performance of GPT-5.4, the best-performing model identified in Section~\ref{subsec:experimental_results}.

The experimental result revealed that the model achieved F1 score of 10.6\% at sentence level and 16.9\% at word level.
The substantial gap between the synthetic and real correction results suggests that real corrections involve additional challenges beyond controlled word-level replacements, such as implicit background knowledge and editorially complex error patterns.
Therefore, the synthetic dataset should be viewed not as a substitute for real corrections, but as a controlled first step for evaluating the intrinsic FED capabilities of LLMs before tackling more complex real-world cases.

\section{Conclusion}

This study analyzed factual errors in human-written text and investigated the span-level performance of LLMs on FED for human-written text.
We utilized correction articles to analyze the tendency of factual errors that are likely to occur in human-written text and uncovered the existence of factual errors characteristic of human-written text.
Based on our analysis, we also devised word replacement methods and synthesized a dataset using these methods.

We also evaluated the performance of recent LLMs on FED at both the sentence-level and word-level.
As a result, we demonstrated that GPT-5.4, as well as fine-tuned models on the FED task, showed a certain performance.
Furthermore, the high detection performance for Unit/Period and Digit suggests that LLMs can detect factual errors using general knowledge.
On the other hand, the best word-level F1 score for detecting factually incorrect spans is still 52\%, indicating that factual error detection is challenging.

\section*{Limitations}

This study has a few limitations, which point to important directions for future work.

\paragraph{Language}

First, we conducted analyses and experiments in Japanese, which limits the generaliability of our results to non-Japanese languages.
However, since this study is not a replication in Japanese of previous studies that have already been conducted in English, we believe that our monolingual work on non-English language provided sufficient contributions.

Leaving that aside, we presume that our distilled taxonomy is relatively language-independent because named entities, antonyms, and numerical and time expressions are observed across various languages.
In addition, correction articles are available in multiple languages;
therefore, our analysis can be extended to non-Japanese languages.

\paragraph{Data}

Second, the constructed synthetic dataset includes some noisy cases that do not result in factually incorrect text.
As shown in our analysis in Section~\ref{sec:basis}, word replacement does not always alter the factual correctness of the original text.
These cases act as noise for evaluation purposes.
To improve data quality, it is necessary to enhance the current filtering method.

\paragraph{Model}

Third, our experiments focus solely on factual error detection without access to external knowledge sources as our motivation is to investigate whether LLMs, which have acquired immense world knowledge through pre-training, can be used for revision support.
To comprehensively evaluate the capability of LLMs to detect factual errors, future work should evaluate the FED task that assumes access to external knowledge.
Examples include databases or retrieval-augmented systems, which are often crucial in on-the-ground editorial work.

In the evaluation of Section \ref{sec:experiments}, the open-weight models used in our fine-tuning experiments are limited to 8B-parameter models.
Evaluating FED performance across a broader range of model scales remains an important direction for future work.

\paragraph{Reproducibility}

Finally, the evaluation data we used in this paper is built upon proprietary newspaper articles, which are not publicly available.
Although this limitation restricts reproducibility, we addressed this concern by constructing a dataset using Japanese articles published on the open-source Wikinews platform\footref{footnote:wikinews} and confirming that the overall performance trends were similar (cf. Appendix~\ref{sec:experiments-wikinews}).
We also released reproduction code,\footref{footnote:github} which we believe ensures reproducibility to a considerable extent.

\section*{Ethical Considerations}

We affirm that our research complies with the ACL Code of Ethics.

\paragraph{Data Acquisition}
The articles of Nikkei Inc. used in this study were obtained through legitimate means without violating the terms of service, bypassing paywalls, or employing scraping techniques.
In particular, we used resources provided by Nikkei Inc., which were accessible to the authors through proper channels.
No personally identifiable information (PII) or sensitive user data was included or used in our dataset construction.

\paragraph{Use of External Tools}
When utilizing publicly available software libraries and pre-trained language models, we have carefully followed their licenses and intended usage policies.
All tools were used for academic research purposes within the scope of what their creators have authorized.

\paragraph{AI Assistance}
We used AI-based writing assistants (e.g., large language models) to support the authors in drafting or proofreading certain English sentences.
In all cases, the generated content was critically reviewed, edited, and verified by the human authors, who take full responsibility for the final version of the manuscript.
No parts of the paper were solely generated by AI without author verification or oversight.

\appendix
\begin{table*}[!t]
    \centering
    \small
    \begin{NiceTabular}{c|l|c}
    \toprule
    Category & \multicolumn{1}{c|}{Example} & Target Word \\
    \midrule
    Named Entity & 
    \begin{minipage}{0.6\textwidth}
    1982年スペイン大会から94年\textbf{\textcolor{red}{ブラジル}}大会までW杯本大会の出場枠は「24」だった。\\
    (From the 1982 tournament in Spain to the 1994 tournament in \textbf{\textcolor{red}{Brazil}}, the number of teams participating in the World Cup finals was ``24''.)
    \end{minipage} &
    \raisebox{-0.3125\height}{\shortstack{米国\\(America)}} \\
    \midrule
    Kanji Misconversion & 
    \begin{minipage}[h]{0.6\textwidth}
    66キロ級の\textbf{\textcolor{red}{安倍一二三}}や73キロ級で2016年リオデジャネイロ五輪覇者の大野将平らが精力的に稽古した。\\
    (Athletes such as \textbf{\textcolor{red}{Hifumi Abe}} in the 66kg class and Shohei Ono, the gold medalist in the 73kg class at the 2016 Rio de Janeiro Olympics, trained vigorously.)
    \end{minipage} &
    \raisebox{-0.3125\height}{\shortstack{阿部一二三\\(Hifumi Abe)}} \\
    \midrule
    Antonym & 
    \begin{minipage}[h]{0.6\textwidth}
    9月の販売価格は32万ドルで、前年同月比3.5\%\textbf{\textcolor{red}{値上がり}}した。\\
    (The sales price in September was \$320,000, a 3.5\% \textbf{\textcolor{red}{increase}} compared to the same month last year.)
    \end{minipage} &
    \raisebox{-0.3125\height}{\shortstack{値下がり\\(decrease)}} \\
    \midrule
    Numerical Value &
    \begin{minipage}[h]{0.6\textwidth}
    地元紙デトロイト・ニュースによると、従業員約4万6000人が参加し、北米の\textbf{\textcolor{red}{29}}工場で操業が止まった。\\
    (According to the local newspaper, The Detroit News, approximately 46,000 employees participated, and operations were halted at \textbf{\textcolor{red}{29}} plants across North America.)
    \end{minipage} &
    31 \\
    \midrule
    Date &
    \begin{minipage}[h]{0.6\textwidth}
    \textbf{\textcolor{red}{2004}}年1月に米アップルが初代「iPhone」を発表してから10年\\
    (It has been 10 years since Apple unveiled the first-generation ``iPhone'' in January \textbf{\textcolor{red}{2004}}.)
    \end{minipage} &
    2007\\
    \midrule
    Digit &
    \begin{minipage}[h]{0.6\textwidth}
    ソフトバンクGは2016年にアームを約\textbf{\textcolor{red}{3億}}円で買収した。\\
    (SoftBank Group acquired ARM for approximately \textbf{\textcolor{red}{300 million}} yen in 2016.)
    \end{minipage} &
    \raisebox{-0.3125\height}{\shortstack{3兆\\(3 trillion)}}\\
    \midrule
    Unit/Period & \begin{minipage}[h]{0.6\textwidth}
    スービック湾は南シナ海に面し、中国が実効支配するスカボロー礁から約200\textbf{\textcolor{red}{マイル}}と近い。\\
    (Subic Bay faces the South China Sea and is approximately 200 \textbf{\textcolor{red}{miles}} from Scarborough Shoal, which is under China's effective control.)
    \end{minipage} &
    \raisebox{-0.3125\height}{\shortstack{キロ\\(km)}} \\
    \midrule
    \end{NiceTabular}
    \caption{
        Examples of automatically generated data for each category.
        The text highlighted in red is a replaced span, and the term ``target word'' represents a word or number before replacement.
    }
    \label{tab:examples_for_fed_data}
\end{table*}

\section{Definitions of Categories}
\label{appendix:definitions_of_categories}
We present the definitions of each category of factual errors reported in correction articles below.
\begin{description}
    \item[Named Entity] Corrections of named entity errors, such as incorrect names of persons, organizations, locations, and so forth.
    \item [Kanji Misconversion] Corrections of homophone errors that arise when an incorrect kanji with the same reading is selected during conversion from hiragana to kanji.
    \item[Synonym] Corrections of synonym errors, i.e., replacing words with ones that have similar meanings.
    \item[Antonym] Corrections of antonym errors, i.e., replacing words with ones that have opposite meanings.
    \item[Other] Corrections of word errors not included in the above four categories, such as factual errors of tense and common nouns.
    \item [Numerical Value] Corrections of number errors to the extent that digits remain the same.
    \item [Date] Corrections of number errors regarding dates.
    \item [Digit] Corrections of number errors involving digit changes.
    \item [Unit/Period] Corrections of numerical errors regarding units or periods.
    \item [Grammatical Error] Corrections of grammatical errors, not factual ones.
    \item [Phrase-Level Error] Corrections of phrase-level errors that cannot be fixed by word replacement alone.
\end{description}

\section{Examples of Constructed Data}\label{appendix:example_data}
Table~\ref{tab:examples_for_fed_data} shows examples of the constructed data described in Section~\ref{subsec:dataset}.

\section{Prompt Template for Filtering}\label{appendix:template}
Figure \ref{fig:prompt_templates} illustrates the prompt templates used in the filtering method of Section \ref{subsec:dataset}.
\begin{figure}[!t]
    \centering
    \begin{screen}
        \begin{minipage}{1.0\linewidth}
            以下の文章中の「\{target\_word\}」は「\{replacement\}」と同義語であるかを判定してください。\\
            (Judge whether the word ``\{target\_word\}'' in the following text is synonymous with ``\{replacement\}''.)\\
            回答は、同義語であれば\{"answer": true\}、そうでなければ\{"answer": false\}というJSON形式で出力してください。\\
            (Output the response in JSON format as \{``answer'': true\} if they are synonymous, and \{``answer'': false\} otherwise.)\\
            \\
            文章: \{text\}
        \end{minipage}
    \end{screen}
    \caption{
        Prompt template for filtering method.
    }
    \label{fig:prompt_templates}
\end{figure}

\begin{table}[!t]
    \centering
    \small
    \begin{NiceTabular}{ll|r}
        \toprule
            \multicolumn{2}{c|}{Parameter} & \multicolumn{1}{c}{Value} \\
        \midrule
            \Block{7-1}{Fine-Tuning} & Max training steps & 1,024 \\
            & Effective batch size & 32 \\
            & Learning rate & 1e-4 \\
            & Optimizer & AdamW \\
        \arrayrulecolor{ruled} \cmidrule{2-3}
            & \Block{2-1}{Scheduler} & cosine schedule \\
            & & with warmup \\
        \cmidrule{2-3}
            & max\_grad\_norm & 0.3 \\
        \arrayrulecolor{black} \midrule
            \Block{9-1}{LoRA} & r & 32 \\
            & lora\_alpha & 16 \\
            & lora\_dropout & 0.05 \\
            & bias & none \\
        \arrayrulecolor{ruled} \cmidrule{2-3}
            & \Block{5-1}{target\_modules} &
            q\_proj, k\_proj,\\
            & & v\_proj, o\_proj,\\
            & & gate\_proj,\\
            & & up\_proj,\\
            & & down\_proj\\
        \arrayrulecolor{black} \midrule
            \Block{2-1}{Quantization} & bits & 4 bit \\
            & quant\_type & nf4 \\
        \bottomrule
    \end{NiceTabular}
    \caption{
      Hyperparameters for fine-tuning of Qwen3-Swallow-8B and Gemma-4-8B.
    }
    \label{tab:qlora_hyperparameters}
\end{table}

\begin{table}[!t]
    \small
    \centering
    \begin{NiceTabular}{l|r|r}
        \toprule
            \multicolumn{1}{c|}{Category} &
            \multicolumn{1}{c|}{Train} &
            \multicolumn{1}{c}{Eval} \\
        \midrule
            Named Entity & 5,458 & 498\\ 
            Kanji Misconversion & 2,000 & 100 \\
            Antonym & 2,000 & 100\\
            Numerical Value & 2,000 & 100\\ 
            Date & 2,000 & 100 \\ 
            Digit & 1,452 & 100 \\ 
            Unit/Period & 431 & 100 \\ 
            no word replacement & 2,841 & 100 \\
        \bottomrule
    \end{NiceTabular}
    \caption{
    Statistics of training and evaluation data generated from Wikinews.
    }
    \label{appendix:wikinews_dataset}
\end{table}

\begin{table*}[!t]
    \small
    \centering
    \begin{NiceTabular}{l|l|rrr|rrr}    
        \toprule
            \multicolumn{1}{c}{\multirow{2}{*}{Model}} &
            \multicolumn{1}{c}{\multirow{2}{*}{Setting}} &
            \multicolumn{3}{c}{Sentence-Level} &
            \multicolumn{3}{c}{Word-Level} \\
            &
            &
            \multicolumn{1}{c}{Prec.} &
            \multicolumn{1}{c}{Rec.} &
            \multicolumn{1}{c}{F1} &
            \multicolumn{1}{c}{Prec.} &
            \multicolumn{1}{c}{Rec.} &
            \multicolumn{1}{c}{F1} \\
        \midrule
            GPT-5.4 &
            Few-Shot ($k = 6$) &
            40.9 &
            52.0 &
            \textbf{45.8} &
            51.2 &
            \textbf{40.2} &
            \textbf{45.0} \\
        \midrule
            \Block{3-1}{Qwen3-Swallow-8B} &
            Few-Shot ($k = 6$) &
            11.0 &
            \textbf{58.5} &
            18.5 &
            63.7 &
            6.0 &
            11.0 \\
            &
            Fine-Tuning (QLoRA) &
            43.0 &
            46.9 &
            44.9 &
            30.0 &
            28.4 &
            29.2 \\
            &
            Fine-Tuning (Full-Parameter) &
            36.3 &
            36.6 &
            36.4 &
            23.8 &
            21.6 &
            22.7 \\
        \midrule
            \Block{3-1}{Gemma-4-8B} &
            Few-Shot ($k = 6$) &
            22.0 &
            19.5 &
            20.7 &
            37.9 &
            8.9 &
            14.4 \\
            &
            Fine-Tuning (QLoRA) &
            \textbf{43.2} &
            38.9 &
            40.9 &
            27.4 &
            21.9 &
            24.3 \\
            &
            Fine-Tuning (Full-Parameter) &
            8.4 &
            6.0 &
            7.0 &
            \textbf{100.0} &
            5.5 &
            10.5 \\
        \bottomrule
    \end{NiceTabular}
    \caption{Performance of detecting factually incorrect spans on the evaluation data generated from Wikinews.}
    \label{tab:wiki_score}
\end{table*}

\begin{table*}[!t]
    \small
    \centering
    \begin{NiceTabular}{l|rrrrrrr}    
        \toprule
            \Block[c]{2-1}{Model} &
            \Block[c]{2-1}{Named\\Entity} &
            \Block[c]{2-1}{Kanji\\Misconversion} &
            \Block[c]{2-1}{Antonym} &
            \Block[c]{2-1}{Numerical\\Value} &
            \Block[c]{2-1}{Date} &
            \Block[c]{2-1}{Digit} &
            \Block[c]{2-1}{Unit/Period} \\ 
            & & & & & & & \\
        \midrule
            GPT-5.4 &
            \textbf{43.7} &
            40.8 &
            \textbf{41.2} &
            \textbf{37.3} &
            \textbf{34.0} &
            \textbf{70.2} &
            57.1 \\ 
            Qwen3-Swallow-8B-QL &
            22.0 &
            \textbf{49.0} &
            33.6 &
            23.5 &
            16.0 &
            60.5 &
            \textbf{57.5} \\
        \bottomrule
    \end{NiceTabular}
    \caption{Word-level F1 scores of detecting factually incorrect spans on the evaluation data generated from Wikinews for each category.}
    \label{tab:wiki_category_score}
\end{table*}

\section{Experiments Using Wikinews}
\label{sec:experiments-wikinews}
For reproducibility, we also constructed a dataset using Japanese articles published on the open-source Wikinews platform\footref{footnote:wikinews} and evaluated the FED capability on it.
Table~\ref{appendix:wikinews_dataset} shows the statistics of the training and evaluation data generated from the dump data of Wikinews from August 1, 2025.
The experimental settings are the same as described in Section \ref{sec:experiments}.

The evaluation results are organized in Table~\ref{tab:wiki_score}.
GPT-5.4 achieved the highest scores in both the sentence-level and word-level settings.
Across all evaluation metrics, fine-tuning performance outperforms few-shot performance for both Qwen3-Swallow-8B and Gemma-4-8B.
Thus, we confirmed that the FED capability also improves by fine-tuning even using Wikinews.

Compared to the evaluation results in Table~\ref{tab:experimental_results}, the performance of GPT-5.4, Qwen3-Swallow-8B-QL, and Gemma-4-8B-QL has decreased.
This indicates that the evaluation data constructed from Wikinews is more challenging than that constructed from Nikkei articles.

The F1 scores for each category are organized in Table~\ref{tab:wiki_category_score}.
For all the models, the performance was high for the Unit/Period and Digit categories.
Therefore, these models possess factual knowledge to detect errors that deviate significantly from facts.
For the Qwen3-Swallow-8B-QL model, its capability to detect errors of the Date category is markedly low, indicating that it is difficult to improve detection performance for the Date category by fine-tuning on synthetic data.

\section{Hyperparameters}
\label{appendix:qlora_hyperparameter}
Table~\ref{tab:qlora_hyperparameters} organizes the hyperparameters for fine-tuning Qwen3-Swallow-8B and Gemma-4-8B.

\section{Full Experimental Results Using Nikkei Articles}
\label{sec:full_experiments_nikkei}

Table~\ref{tab:full_experimental_results} shows the performance of detecting factually incorrect spans on the evaluation data generated from Nikkei articles.
We attempted full-parameter fine-tuning, yielding lower performance than QLoRA.
In particular, the full-parameter fine-tuned Gemma-4-8B model appears to learn a shortcut, frequently responding to most test cases with ``There are no factual errors.'', which is likely to cause its poor performance.

\begin{table*}[!t]
    \small
    \centering
    \begin{NiceTabular}{l|l|rrr|rrr}    
    \toprule
    \multicolumn{1}{c}{\multirow{2}{*}{Model}} & \multicolumn{1}{c}{\multirow{2}{*}{Setting}} & \multicolumn{3}{c}{Sentence-Level} & \multicolumn{3}{c}{Word-Level} \\
     & &
     \multicolumn{1}{c}{Prec.} & \multicolumn{1}{c}{Rec.} & \multicolumn{1}{c}{F1} &
     \multicolumn{1}{c}{Prec.} & \multicolumn{1}{c}{Rec.} & \multicolumn{1}{c}{F1} \\
    \midrule
    GPT-5.4 & Few-Shot ($k = 6$) & \textbf{41.7} & 46.1 & \textbf{43.8} & 78.7 & \textbf{38.8} & \textbf{52.0} \\
    \midrule
    \Block{3-1}{Qwen3-Swallow-8B} & Few-Shot ($k = 6$) & 9.0 & \textbf{48.5} & 15.2 & 45.0 & 6.2 & 10.9 \\
     & Fine-Tuning (QLoRA) & 41.3 & 36.4 & 38.7 & 49.2 & 28.7 & 36.2 \\
     & Fine-Tuning (Full-Parameter) & 31.9 & 27.6 & 29.6 & 36.2 & 20.4 & 26.1 \\
    \midrule
    \Block{3-1}{Gemma-4-8B} & Few-Shot ($k = 6$) & 18.4 & 15.5 & 16.8 & 51.6 & 8.6 & 14.7 \\
     & Fine-Tuning (QLoRA) & 32.1 & 25.9 & 28.7 & 56.7 & 21.1 & 30.8 \\
     & Fine-Tuning (Full-Parameter) & 9.6 & 7.3 & 8.3 & \textbf{98.3} & 7.0 & 13.0 \\
    \bottomrule
    \end{NiceTabular}
    \caption{
        Performance of detecting factually incorrect spans on the evaluation data generated from Nikkei articles.
    }
    \label{tab:full_experimental_results}
\end{table*}

\begin{table*}[!t]
    \centering
    \small
    \begin{NiceTabular}{c|l|c}
        \toprule
            Difficulty & \multicolumn{1}{c|}{Example} & Target Word \\
        \midrule
            Noise 1 & 
            \begin{minipage}{0.6\textwidth}
                エアロダイングループはアジア太平洋地域から、中東、\textbf{\textcolor{red}{北米}}、ヨーロッパなど数十カ国で事業を展開している。\\
                (The Aerodyne Group operates in dozens of countries, from Asia Pacific to the Middle East, \textbf{\textcolor{red}{North America}}, and Europe.)
            \end{minipage} &
            \raisebox{-0.3125\height}{\shortstack{アメリカ大陸\\(Americas)}} \\
        \midrule
            Noise 2 & 
            \begin{minipage}[h]{0.6\textwidth}
                神奈川県内の上場企業119社の決算を集計したところ、4～\textbf{\textcolor{red}{4}}月期の経常利益は前年同期比63\%減の1003億円だった。\\
                (According to a compilation of financial results from 119 listed companies in Kanagawa Prefecture, recurring profit for the from April to \textbf{\textcolor{red}{April}} period was down 63\% from the same period last year to 100.3 billion yen.)
            \end{minipage} &
            \raisebox{-0.3125\height}{\shortstack{9\\(September)}} \\
        \midrule
            Difficulty 1 & 
            \begin{minipage}[h]{0.6\textwidth}
                社長に就いた15年6月期に306店舗だったグループは、19年6月期に693店舗まで\textbf{\textcolor{red}{減少}}。\\
                (The group, which had 306 stores in the fiscal year ending June 2015 when he took over as president, \textbf{\textcolor{red}{decreased}} to 693 stores in the fiscal year ending June 2019.)
            \end{minipage} &
            \raisebox{-0.3125\height}{\shortstack{増加\\(increased)}} \\
        \midrule
            Difficulty 2 &
            \begin{minipage}[h]{0.6\textwidth}
                \textbf{\textcolor{red}{オバマ}}米大統領のプライベートバンカーを長年務めてきたローズマリー・ブラブリック氏が、勤務先のドイツ銀行を辞職した。\\
                (Rosemary Brabrick, a longtime private banker to U.S. President \textbf{\textcolor{red}{Obama}}, has resigned from Deutsche Bank, where she worked)
            \end{minipage} &
            \raisebox{-0.3125\height}{\shortstack{トランプ\\(Trump)}} \\
        \midrule
            Difficulty 3 &
            \begin{minipage}[h]{0.6\textwidth}
                米ワシントン・ポスト紙によると、調査会社の調べでEUは米国よりファイザー製で約\textbf{\textcolor{red}{31\%}}。\\
                (According to the Washington Post, a research firm found that the EU is about \textbf{\textcolor{red}{31\%}} more Pfizer-made than the U.S.)
            \end{minipage} &
            24\% \\
        \bottomrule
    \end{NiceTabular}
    \caption{
        Examples for each detection difficulty.
        The text highlighted in red is a factually incorrect span.
        The term ``target word'' represents a word or number before replacement.
    }
    \label{tab:examples_for_ddt}
\end{table*}

\section{Examples for Each Detection Difficulty}
\label{appendix:examples_for_ddt}
Table \ref{tab:examples_for_ddt} shows examples for each detection difficulty defined in Section~\ref{sec:basis}.

\section{Generalizability of Taxonomy to More Recent Corrections}
To examine whether the similar trend can be observed in more recent correction articles, we manually classified correction articles published \emph{between January 2024 and July 2026} into the same taxonomy as that in Table \ref{tab:category}.

The classification results are organized in Table \ref{tab:category_2}.
We found that our taxonomy generalized well to errors reported in more recent correction articles.

\begin{table}[!t]
    \centering
    \small
    \begin{NiceTabular}{c|c|c}
        \toprule
            Supercategory &
            Category &
            Count \\
        \midrule
            \Block{5-1}{Word Error} & 
            \raisebox{-0.3125\height}{\shortstack{Named\\Entity}} &
            33 \\
        \cmidrule{2-3}
             &
            \raisebox{-0.3125\height}{\shortstack{Kanji\\Misconversion$^{\dagger}$}} & 
            9 \\
        \cmidrule{2-3}
            &
            Synonym & 
            7 \\
        \cmidrule{2-3}
             &
            Antonym & 
            9 \\
        \cmidrule{2-3}
             &
            Other & 
            3 \\
        \midrule
            \Block{4-1}{Number Error} &
            \raisebox{-0.3125\height}{\shortstack{Numerical\\Value}} &
            20 \\
        \cmidrule{2-3}
             &
            Date & 
            10 \\
        \cmidrule{2-3}
             &
            Digit$^{\dagger}$ & 
            9 \\
        \cmidrule{2-3}
             &
            Unit/Period$^{\dagger}$ & 
            6 \\
        \midrule
        \midrule
            \Block{1-2}{Grammatical Error} & & 
            3 \\
        \midrule
            \Block{1-2}{Phrase-Level Error} & & 
            13 \\
        \bottomrule
    \end{NiceTabular}
    \caption{
        Classification result of factual errors and their corrections reported in 122 Japanese correction articles.
        $\dagger$ denotes the category of factual errors that is characteristic of human-written text.
    }
    \label{tab:category_2}
\end{table}

\section{Effect of Few-Shot Configuration}
To investigate how performance changes with the number of shots, we evaluated the zero-shot, 5-shot, and 10-shot performance of GPT-5.4. 
We evaluated the models using the F1 scores at both word-level and sentence-level.

Table \ref{appendix:few-shot_eval} shows the experimental results.
Since the performance gap between the zero-shot, 5-shot, and 10-shot settings is marginal, we presume that the experimental results would not be significantly affected by the few-shot configuration.

\begin{table}[!t]
    \small
    \centering
    \begin{NiceTabular}{l|r|r}
        \toprule
            \multicolumn{1}{c|}{Setting} &
            \multicolumn{1}{c|}{Sentence-Level} &
            \multicolumn{1}{c}{Word-Level} \\
        \midrule
            Zero-Shot & 42.9 & 47.6 \\ 
            Few-Shot ($k = 5$) & 44.2 & 51.5 \\
            Few-Shot ($k = 6$) & 43.8 & 52.0 \\
            Few-Shot ($k = 10$) & 44.8  & 51.2 \\ 
        \bottomrule
    \end{NiceTabular}
    \caption{
    Performance of detecting factually incorrect spans for each few-shot configuration.
    All the results were obtained using GPT-5.4.
    }
    \label{appendix:few-shot_eval}
\end{table}

\end{document}